\documentclass[review]{elsarticle}

\usepackage{color}
\usepackage{lineno,hyperref}
\usepackage{verbatim}
\usepackage{amssymb,amsmath}
\usepackage{algorithm,algorithmic}
\usepackage{graphicx}
\usepackage{xcolor}
\usepackage{epstopdf}
\definecolor{blue}{rgb}{0.0, 0.46, 0.37}

\journal{Journal of \LaTeX\ Templates}

\begin{document}

\begin{frontmatter}

\title{Sequential vessel segmentation via deep channel attention network}

\author[add1]{Dongdong Hao\corref{co-firstauthors}}
\author[add2]{Song Ding\corref{co-firstauthors}}
\author[add3]{Linwei Qiu}
\author[add4]{Yisong Lv}
\author[add5]{Baowei Fei}
\author[add6]{Yueqi Zhu}
\author[add1]{Binjie Qin\corref{mycorrespondingauthor}}
\ead{bjqin@sjtu.edu.cn}

\cortext[co-firstauthors]{The co‐first authors contributed equally to this work}
\cortext[mycorrespondingauthor]{Corresponding author}

\address[add1]{School of Biomedical Engineering, Shanghai Jiao Tong University, Shanghai 200240, China}
\address[add2]{Department of Cardiology, Ren Ji Hospital, School of Medicine, Shanghai Jiao Tong University, Shanghai 200127, China.}
\address[add3]{School of Astronautics, Beihang University, Beijing 100191, China}
\address[add4]{School of Continuing Education, Shanghai Jiao Tong University, Shanghai 200240, China}
\address[add5]{Department of Bioengineering, Erik Jonsson School of Engineering and Computer Science, University of Texas at Dallas, Richardson, TX 75080, USA}
\address[add6]{Department of Radiology, Shanghai Jiao Tong University Affiliated Sixth People’s Hospital, Shanghai Jiao Tong University, 600 Yi Shan Road, Shanghai 200233, China.}

\begin{abstract}
Accurately segmenting contrast-filled vessels from X-ray coronary angiography (XCA) image sequence is an essential step for the diagnosis and therapy of coronary artery disease. However, developing automatic vessel segmentation is particularly challenging due to the overlapping structures, low contrast and the presence of complex and dynamic background artifacts in XCA images. This paper develops a novel encoder-decoder deep network architecture which exploits the several contextual frames of 2D+t sequential images in a sliding window centered at current frame to segment 2D vessel masks from the current frame. The architecture is equipped with temporal-spatial feature extraction in encoder stage, feature fusion in skip connection layers and channel attention mechanism in decoder stage. In the encoder stage, a series of 3D convolutional layers are employed to hierarchically extract temporal-spatial features. Skip connection layers subsequently fuse the temporal-spatial feature maps and deliver them to the corresponding decoder stages. To efficiently discriminate vessel features from the complex and noisy backgrounds in the XCA images, the decoder stage effectively utilizes channel attention blocks to refine the intermediate feature maps from skip connection layers for subsequently decoding the refined features in 2D ways to produce the segmented vessel masks. Furthermore, Dice loss function is implemented to train the proposed deep network in order to tackle the class imbalance problem in the XCA data due to the wide distribution of complex background artifacts. Extensive experiments by comparing our method with other state-of-the-art algorithms demonstrate the proposed method's superior performance over other methods in terms of the quantitative metrics and visual validation. To facilitate the reproductive research in XCA community, we publically release our dataset and source codes at \url{https://github.com/Binjie-Qin/SVS-net}.
\end{abstract}


\end{frontmatter}


\section{Introduction}
\label{sec:intro}
\subsection{Motivation}
Nowadays, cardiovascular diseases have seriously threatened more and more people’s health \citep{Townsend20163232}. Percutaneous coronary intervention, as the minimally invasive method for cardiovascular disease treatment, has been widely adopted in the clinic. During this intervention, contrast agents are injected into the vessels through one catheter and then X-ray coronary angiography (XCA\footnote{The notations used in this paper are listed in Table \ref{notation}.}) is employed to help surgeons navigate the catheters \citep{JIN2017653,Albarqouni2017444,J.20091563}. With the help of the contrast-enhanced images, doctors can diagnose the coronary artery disease and evaluate therapeutic effects. It is important to accurately and quickly segment vessels from XCA data for the diagnosis and intervention of cardiovascular diseases. 

Although vessel segmentation has always been a hot spot due to its significance and complexity in clinical practice \citep{soares2006retinal, KERKENI201649, jin2018low}, coronary artery vessel segmentation still remains highly challenging because of the poor visual quality of XCA, which is caused by the low contrast and high Poisson noise of low dose X-ray imaging, overlap of background structures and foreground vessels, complex motion patterns and disturbance of spatially distributed noisy artifacts. Currently most 3D/2D vessel segmentation methods are proposed to segment the vessels from computed tomography angiography, magnetic resonance angiography and a single 2D image \citep{jin2018low,MOCCIA201871}, of which there is no serious disturbance from the noise and overlapping background structures. To obtain the vascular structures from XCA, a few computer vision and machine learning related methods have been developed. Kerkeni \textit{et al.} \citep{KERKENI201649} propose an iterative region growing algorithm to integrate both vesselness and direction information in the multi-scale space. However, it fails to recognize thin and peripheral vessel in the low contrast XCA images. Jin \textit{et al.} \citep{jin2018low} extract the contrast-filled vessels via robust principal component analysis and combine both local and global threshold to refine vessel segmentation mask \citep{unberath2017consistency} but with some residuals remained around vessel regions. Felfelian \textit{et al.} \citep{felfelian2016vessel} detect coronary artery regions of interest based on Hessian filter and identify vessel pixels by flux flow measurements. Nevertheless, a series of postprocessing should be performed to improve the robustness and accuracy of segmentation mask. 
\begin{table}
	\caption{The main notations used in this paper.}
		\label{notation}       
		\begin{tabular}{ l | p{5.4cm}  }
			\hline\noalign{\smallskip}
			Notation &  Explanation \\
			\noalign{\smallskip}\hline\noalign{\smallskip}
			XCA &  X-ray coronary angiography\\
			SVS-net & sequential vessel segmentation deep network\\
			CRF &  conditional random field\\
			FFO &  feature fusion operation\\
			CAB &  channel attention block\\
			Conv3D & 3D convolution\\
			Block3D& 3D residual convolutional block\\
			Conv2D& 2D convolution\\
			Block2D& 2D residual convolutional block\\
			BN& batch normalization\\
			ReLu& rectified linear unit\\
			sigmoid& sigmoid activation function\\
			ks& convolutional kernel size\\
			LF  &  low-stage feature maps\\
			HF  &  high-stage feature maps\\
			DR  &  detection rate\\
			P   &  precision \\
			CE  &  cross entropy \\
			GVEs&  gland volume errors\\
			\noalign{\smallskip}\hline
	\end{tabular}
\end{table} 

With the development of neural network-based deep learning, Nasr-Esfahani \textit{et al.} \citep{nasr2016vessel} use convolutional neural network (CNN) with fully-connected layers to perform vessel segmentation, which overlooks the structure information and temporal correlation in XCA sequence images. To alleviate these issues, Fan \textit{et al.} \citep{fan2018multichannel} develop a multichannel fully convolutional neural network with live image and corresponding dense matching mask image inputted to the network. However, it should collect corresponding mask images and perform dense matching in advance to segment vessel structure from live images, which is not practical in clinical applications. Most of XCA segmentation algorithms are dependent on the pixels of local windows in a single frame of XCA sequences, so that they waste lots of temporal-spatial contextual information in XCA sequences, which can be important to infer whether the pixels belongs to the foreground vessel regions or not.  Although current vessel segmentation \citep{MOCCIA201871} methods have made great progress in segmentation accuracy, they are still inefficient in the large dynamic datasets from the complex XCA sequences with many noisy and overlapped background artifacts. 

To design a robust and efficient XCA segmentation algorithm for clinical applications, we should have a good knowledge about the XCA images' characteristics. Usually, with the illumination of X-rays at specific angle or direction, various 3D anatomical structures such as vessels, lungs, spines, diaphragms and bones are projected along definite path and displayed as overlapped 2D structures on the X-ray angiogram plane. To simplify description, we straightly identify vessels as foreground and regard other overlapped structures as background. Low dose radiopaque contrast agents are primarily injected into angiocarpy to enhance the visibility of vessels in XCA images. Even so, the vessels in XCA images are still of poor visibility due to the following factors: (1) The projection onto 2D plane causes overlap of adjacent tissues. Therefore, foreground vessel regions are badly disturbed by respiratory motion\citep{blondel2006reconstruction}. Moreover, it is very difficult to differentiate foreground vessels from background due to the low intensity contrast between the vessels and the background in low-dose X-ray imaging \citep{xia2019vessel}; (2) Vessels usually have plenty of branches. Radiopaque contrast agents flow at different speeds in each branches. As a result, different vessels branches vary in gray values and some vessel regions cannot be clearly visible in the same time\citep{QIN201938};  (3) The spatially distributed Poisson noises \citep{Zhu2013Reducing} caused by low-dose X-ray imaging reduce the SNR between the signals and noise. The noisy background artifacts and foreground vessels have different motion patterns, so that these noisy and dynamic structures severely interfere with the feature extraction and classification for vessel segmentation. All abovementioned difficulties determine that sequential vessel segmentation from XCA image sequences is a highly challenging task. 

Hierarchical deep CNN features have proven incredibly effective at a wide range of image classification and image segmentation tasks. The generic deep CNN feature extractor trained for general purpose image segmentation often perform surprising well for novel segmentation tasks without doing any fine-tuning beyond training a linear classifier \citep{chen2018deeplab,ronneberger2015u}. This success is often explained by the built-in invariance of deep CNN features to local image transformation and the insensitivity of deep CNN features to shading, low-contrast, etc. We might hope that these invariances would prove useful in our challenging setting of sequential vessel segmentation. However, our problem differs in that we need to segment sequential foreground vessels from the noisy and overlapped background with similar appearances rather than simply training a k-way classifier. To overcome all the mentioned issues of XCA segmentation by deep network, we give the following specific considerations: 

(1) although unsupervised learning or weakly-supervised learning \citep{kallenberg2016unsupervised,huang2016unsupervised} with deep CNN features have developed a lot, they still fail to obtain competitive performance compared with supervised ones. That is because supervised learning introduces straight priors to guide the learning process. In view of the requirement of segmentation accuracy, we adopt supervised learning strategy. As a data-driven method, supervised deep learning depends on large annotated training datasets to ensure excellent performance especially for the video related tasks. Unfortunately, there is no readily available public dataset for vessel segmentation from XCA sequence. To this end, we have collected many XCA sequences from our university-affiliated hospitals and employed several clinical experts to annotate vessel label so that we can set up ground truth for vessel segmentation.

(2) Recent approaches have concentrated on some but not all the fore-mentioned issues and try to make use of temporal information for vessel segmentation. They either take adjacent multiple frames in a sliding window centered at current frame as whole input straightly \citep{hao2018vessel} or make pre-matching to generate segmentation mask \citep{fan2018multichannel,khanmohammadi2017segmentation}. The former indeed introduces not only temporal information but also much disturbances; the latter needs extra dense image matching which is time consuming and incorrect especially for the low contrast images. Recent video segmentation methods explore how to properly utilize the temporal information in the sequential images, \textit{i.e.}, estimate optical flow\citep{Sun_2018_CVPR,rashed2019optical} for modeling the motion among adjacent frames, apply convolutional LSTMs \citep{pfeuffer2019semantic,pfeuffer2019separable} to learn long short-term dependencies in video sequence. They learn effective temporal-spatial consistent features in natural scene image, however they may exist data matching errors \citep{simoncelli1991probability} when applied in the noisy, low-contrast, and blurry XCA medical images. Therefore, how to design an effective network that can learn proper temporal-spatial vessel features from the noisy background will be most important for our work. 

(3) The class imbalance problem caused by the imbalance ratio between the number of foreground vessel pixels and background pixels typically lies in the challenging vessel segmentation tasks and must be well treated to boost the vessel segmentation. Current methods partly addressed this issue by weighted cross entropy \citep{lim2018learning} or proper training patches selection strategy \citep{nasr2016vessel,yan2018joint}. However, they failed to completely solve that imbalance problem. Inspired by the work in \citep{ambrosini2017fully,8328863}, we utilize Dice loss function to guide the network learn balanced information representation between foreground vessel and background pixels. In addition, the deeper the network, the stronger is the representation capacity. However, the optimization of deep network structure is extremely difficult due to the common problem of gradient vanishing and gradient explosion in deep network. We have properly integrated residual blocks \citep{he2016deep,xie2017aggregated,szegedy2017inception} into our vessel segmentation deep network to alleviate the above problem.

In summary, this work has the following contributions: 

1) We propose an encoder-decoder-based sequential vessel segmentation deep network architecture called \textbf{SVS-net} that acquires the temporal-spatial information from the several contextual frames in a sliding window centered at current frame to segment the 2D vessel masks of the current frame in XCA sequence: i) In encoder network, temporal-spatial vessel features from the complex and noisy background artifacts are extracted in 3D (2D+t) manners; ii) The extracted features are then fused along temporal axis in the skip connection layers, which transform the contextual 3D temporal-spatial feature maps into 2D spatial feature maps for the segmentation of current frame. This 3D-2D fusion introduces dimension reduction to further help reduce the subsequent calculation burden and trainable parameters; iii) Finally, the decoder network efficiently integrates the fused temporal-spatial information in XCA image sequence by feature refinement and subsequently decodes the refined features in 2D ways to produce the segmented vessel masks. Specifically, a channel attention mechanism is implemented in channel attention blocks (CABs) to refine the fused temporal-spatial features by adaptively highlighting and learning the discriminative vessel features from the noisy background artifacts via weighting the feature maps. To the best of our knowledge, it is the first time to apply channel attention mechanism in taking both temporal and spatial information into the deep sequential vessel segmentation architecture. Moreover, the proposed SVS-net can be trained in an end-to-end way. 
 
2) We publically release a XCA database with ground truth annotation. The lack of XCA data with annotated label impedes the further exploration on XCA related researches such as vessel segmentation and vessel recovery in deep learning community. Therefore, we established database to promote these studies with detailed data description in the method section of this paper.

3) We employ Dice loss function in deep network to alleviate the severe class unbalance problem in sequential vessel segmentation and validate its significance when compared with binary cross entropy. We have evaluated the effectiveness of 3D temporal-spatial features and CABs used in the proposed SVS-net by comparing them with 2D counterparts and other state-of-the-art methods. Extensive experiments have verified SVS-net's superior performance over other algorithms.

\subsection{Related works}\label{sec-related}
This section introduces the recent works related to vessel segmentation. Vessel segmentation algorithms can be simply divided into two categories: traditional segmentation methods and deep learning-based methods. Recent traditional methods and deep learning-based methods are summarized in this section respectively.

\subsubsection {Traditional vessel segmentation methods}
Various traditional approaches have emerged in the past decades, including filtering based methods \citep{soares2006retinal,MOCCIA201871,frangi1998multiscale,chaudhuri1989detection}, tracking based algorithms \citep{staal2004ridge,kumar2010radon}, and model-based methods \citep{chen2019minimal,dehkordi2014local,law2009efficient}. Filtering-based methods develop specific filters convolving with the original images to enhance the tubular structures \citep{MOCCIA201871,frangi1998multiscale,chaudhuri1989detection}. In \citep{chaudhuri1989detection}, the intensity profile of the vessel was approximately modeled as a Gaussian shaped curve and then 12 different matched filtering templates are utilized to search for the latent vessel segments along different directions. Frangi \textit{et al.} \citep{frangi1998multiscale} propose a common vesselness enhancement technique, where the second order derivative is calculated to form Hessian matrices and the corresponding eigenvalues are analyzed. 

Different from above filtering based approaches, another classes of filters are developed to extract vessel features, such as the ridges feature, the Radon-like features and Gabor wavelet features \citep{staal2004ridge,kumar2010radon,soares2006retinal} and construct pixel-wise vessel feature descriptors for classification. Although these methods enhance vessel structure to some degree, they are executed with high time complexity for the pixel-wise manipulation. Besides, they usually serve as the preprocessing step and further postprocessing like threshold methods and morphology operation should be utilized to construct final refined vessel masks. 

In regard to tracking based segmentation methods, the initial seed points on the vessel edges are chosen firstly and then the tracking process starts under the guidance of image-derived constraints. The tracking algorithms vary from each other according to the different definition of the tracking constraints. For example, Makowski \textit{et al.} \citep {makowski2002two} employ two-phase based method during vessel extraction, which use balloon segmentation and snake segmentation, respectively. Recursive tracking \citep{carrillo2007recursive} is performed by accumulating pixels on the basis of a cluster algorithm with geometry and intensity constraints, while level set evolution \citep{manniesing2007vessel} is employed to track the vessel axis with the evolution process being guided by imposing shape constraints on the skeleton topology. However, these tracking based methods fail to segment out small vessels from the complex and overlapped noisy background, and human intervention is needed to set and adjust the algorithms' parameters. 

Usually, model-based methods are designed on the basis of specific shapes and appearance of the interested structures \citep{MOCCIA201871,chen2019minimal} contained in the images. There are mainly three categories including parametric model, deformable model-based segmentation, and statistic model. We can refer to \citep{MOCCIA201871} for detailed introduction. The model-based methods remain many challenges on detecting small vessels, finding out right parameters to fit the model, and recognizing abnormalities consisted in the diseased vessels. Overall, above traditional segmentation methods require professional knowledge to elaborately construct feature engineering and the complex processing procedures and their segmentation accuracy and real time performance still need to be improved.  

\subsubsection {Deep learning-based methods}
Compared with traditional segmentation methods, deep learning ones automatically learn proper feature representation and perform better on generalization capacity as well as inference speed. Consequently, deep learning methods can earn a top rank in many computer vision fields including segmentation, detection, classification and so on \citep{voulodimos2018deep,sakkos2019illumination}. 

Recently, CNN-based methods have been broadly applied to medical image segmentation such as retinal vessel segmentation \citep{yan2018joint,de2018clinically}. Generally, the works in \citep{liskowski2016segmenting,nasr2016vessel} treat the retinal vessel segmentation task as binary classification, in which a typical classification network containing several stacked convolutional layers and three fully-connected layers are adopted. To alleviate the limitation of annotated data, a patch-based learning strategy is implemented. However, there exists several problems: (1) The limited size of patch means a limited receptive field, which fails to provide sufficient contextual information for accurate segmentation. Fusing predictions of all patches in the image to form the final vessel mask needs to run the whole network many times and is very time consuming. (2) Fully-connected layers function as feature weighting and fuse both local and global information from feature space to label space. However, they contain almost 80\% the parameters of the whole network, which may result in overfitting \citep{Wen2019,Ruder2018}. (3) Due to the localization requirement, fully-connected layers overlook the spatially structured features that are significant for segmentation tasks. Furthermore, the usage of fully-connected layers sets a limit on the network's input size. 

To deal with these inherent problems, fully convolutional network (FCN) \citep{maninis2016deep, dasgupta2017fully} is proposed to replace fully-connected network in segmentation tasks. Recently, a FCN based on encoder-decoder architecture is introduced in \citep{fan2018multichannel}, which adopts a two-channel input strategy and largely depends on the pre-matching between the two-channel inputs. Mo \textit{et al.} \citep{mo2017multi} combine some intermediate layers’ outputs and fuse hierarchical features together to set up the final segmentation map. Similarly, a deeply supervised multi-level and multi-scale network with short connections is utilized to ease the gradient back propagation for retinal vessel segmentation \citep{guo2018deeply}. However, proper feature fusing weights should be carefully set. Fu \textit{et al.} \citep{fu2016retinal} have modeled the retinal vessel segmentation as a pixel‐level classification based on modified FCN. Unfortunately, the lack of smoothness constraint and the limited receptive fields in FCN result in false positive (spurious) regions in segmentation output. Therefore, conditional random field (CRF) formulating long-range interactions between pixels is employed to refine the coarse vessel maps \citep{hu2018retinal}. However, most vessel segmentation algorithms are proposed for solely segmenting vessels from 3D and/or 2D vessel images, which are not appropriate to confront the poor visual quality as well as complex and dynamic background artifacts in sequential vessel segmentation of XCA sequences. 

To focus on the most salient features and suppress the less relevant artifacts simultaneously during learning, attention mechanism equipped within deep learning network \citep{chen2017sca, hu2018squeeze, jetley2018learn} is widely adopted for various tasks including image classification \citep{wang2017residual, peng2018object, schlemper2019attention}, image segmentation \citep{yu2018learning, schlemper2019attention,li2019attention,kearney2019attention} and object detection \citep{li2016deepsaliency,li2018contrast,fu2018refinet}. Attention mechanism is derived from the study of human visual mechanisms, with which people usually pay more attention to the most salient information while neglect some trivials. The key idea of attention mechanism lies in properly generating attention maps to weight feature maps which are extracted by convolutional layers. Zhou \textit{et al.} \citep{zhou2016learning} use fully convolutional networks and utilize global average pooling to generate attention maps. Hu \textit{et al.} \citep{hu2018squeeze} and Yu \textit{et al.} \citep{yu2018learning} have proposed channel attention mechanism to obtain weight vectors by modeling the channel-wise relationship between different feature maps. Chen \textit{et al.} \citep{chen2017sca} integrate both spatial and channel-wize attention in CNN for image captioning. However, to the best of our knowledge, there are no deep network utilizing the channel attention mechanism to extract most salient vessel features from complex and dynamic background artifacts in spatial-temporal contexts for XCA vessel segmentation.
\begin{figure*}
	\centering
	\includegraphics[width=1.2\linewidth]{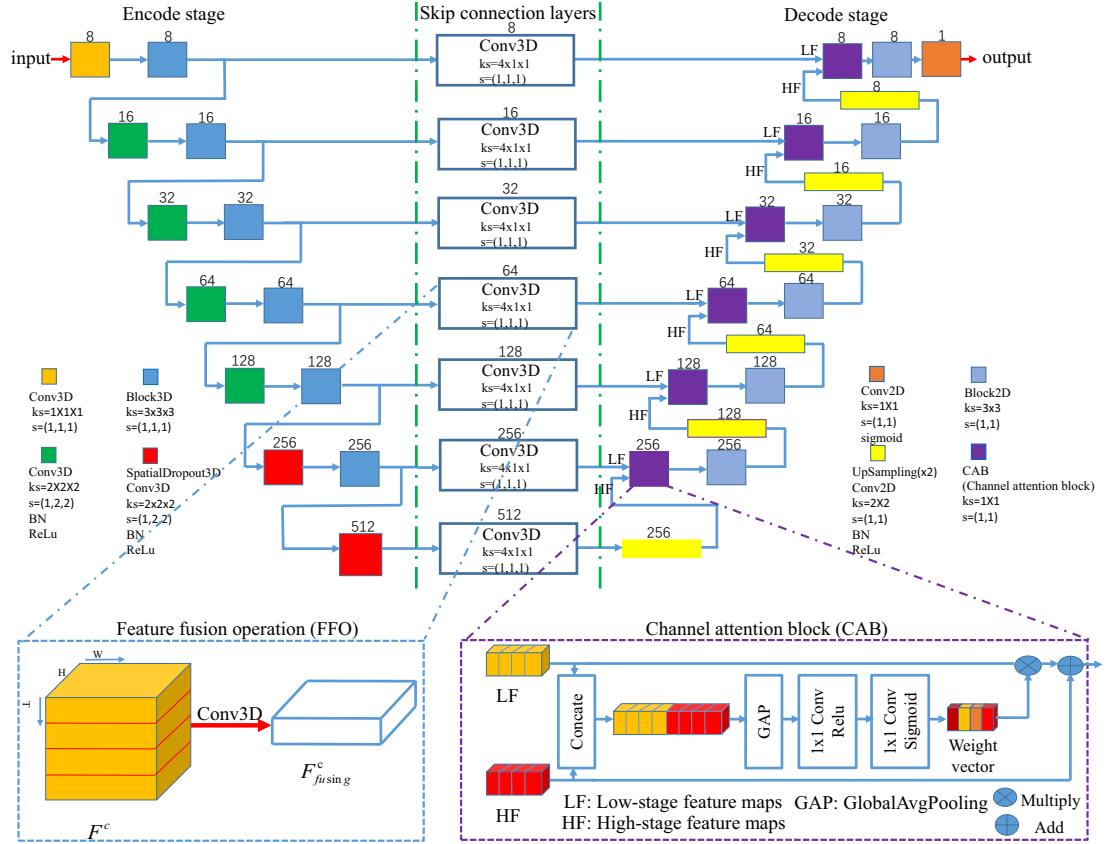}
	\caption{The proposed network architecture is based on U-net with the encoder network extracting 3D feature from the input sequence and the decoder network learning the salient feature via upsampling and operation of CAB, between the encoder and decoder network is the skip connection layers with FFO. The number 8, 16, 32,..., above each block denoting the number of output channels for that block. Convolutional kernel sizes and strides (s: strides) for each block are given in legend. In the FFO and CAB in the bottom, the $ F \in \mathbb{R}^{C\times T \times H \times W}$ denotes the temporal-spatial feature maps, $C$ denotes channel axis, T denotes temporal axis, $H$ denotes height axis, $W$ denotes width axis, $F^{c} \in \mathbb{R}^{T\times H \times W}$: the $c$th channel of temporal-spatial feature maps. $F^{c}_{fusing} \in \mathbb{R}^{H \times W} $ denotes the $c$th channel of fused temporal-spatial feature map through Conv3D with kernel size $4\times1\times1$ and strides (1,1,1).}
	\label{Fig1}
\end{figure*}

\section{Methods}\label{sec-method}
\subsection{Overview}
The architecture is equipped with temporal-spatial feature extraction in encoder stage, feature fusion operation (FFO) in skip connection layers and CAB in decoder stage. In the encoder stage, a series of 3D convolutional layers are employed to hierarchically extract temporal-spatial features. Skip connection layers subsequently fuse the temporal-spatial feature maps and deliver them to the corresponding decoder stages. To learn discriminative feature representation and suppress the complex and noisy artifacts in the XCA images, the decoder stage effectively utilizes CAB to refine the intermediate feature maps from skip connection layers.  

In the proposed SVS-net, (1) we introduce 3D residual blocks (see Fig. \ref{Fig1}) to extract multi-scale temporal-spatial features while ease network optimization in feature encoder stage; (2) these 3D features are integrated by the skip connection layers (see Fig. \ref{Fig1}), which fuse the temporal-spatial 3D feature maps along temporal axis and generate the fused 2D spatial feature maps. Through the fusion in the left bottom of Fig. \ref{Fig1}), the feature maps' dimension mismatch problems between the 3D encoder stage and the 2D decoder stage are solved and the computation cost is also reduced; (3) the fused features are passed to CAB (see the right bottom of Fig. \ref{Fig1}) to refine the vessel features from the noisy background and then transmitted to the decoder stage. (4) Furthermore, Dice loss function is implemented to train the proposed deep network in order to tackle the class imbalance problem in the XCA data due to the imbalanced  ratio between background pixels and foreground pixels. CAB in the decoder stage and FFO in the skip connection layers used in the proposed architecture are also displayed in the bottom of Fig. \ref{Fig1}.

In the following part of this section, we illustrate the architecture and its training setup in detail. Data augmentation methods and loss function are also introduced in this section.
 
\subsection {Experimental Setup} \label{sec2.2}
The XCA image sequence consists of a set of frames ($F_{1}$,$F_{2}$,...$F_{n}$). Each frame $F_{i}$ corresponds to a binary probability map $Y_{i}$ where the value of the foreground vessel pixels is 1 and the other background regions is 0. For intuitive perspective, single frame fails to provide enough contextual information to infer one pixel belonging to foreground or background because of the low contrast of intensity and the similar appearance between the foreground and background. Among successive frames, contrast-filled vessel regions move fast and consistently through the contiguous frames and the noisy and dynamic background artifacts fluctuate synchronously along with human breathing and heart beating. Therefore, these consistent contexts can serve as the auxiliary temporal-spatial information to accurately identify vessels from background. In this work, we experientially use successive 4 frames (\textit{i.e.}, $F_{i-2},F_{i-1}$,$F_{i}$,$F_{i+1}$) as input to generate predicted probability map (\textit{i.e.}, $P_{i}$) with considering that too many frames will increase the burden of memory and calculation. Furthermore, due to the salient motion disturbances introduced by heart beating and breathing in a relative long period, too many frames will result in big differences of the vessel's shapes and positions between the first and last frame, causing the temporal-spatial contexts turning into misleading information.
 \begin{figure*}
 	\centering
 	\includegraphics[width=1.2\linewidth]{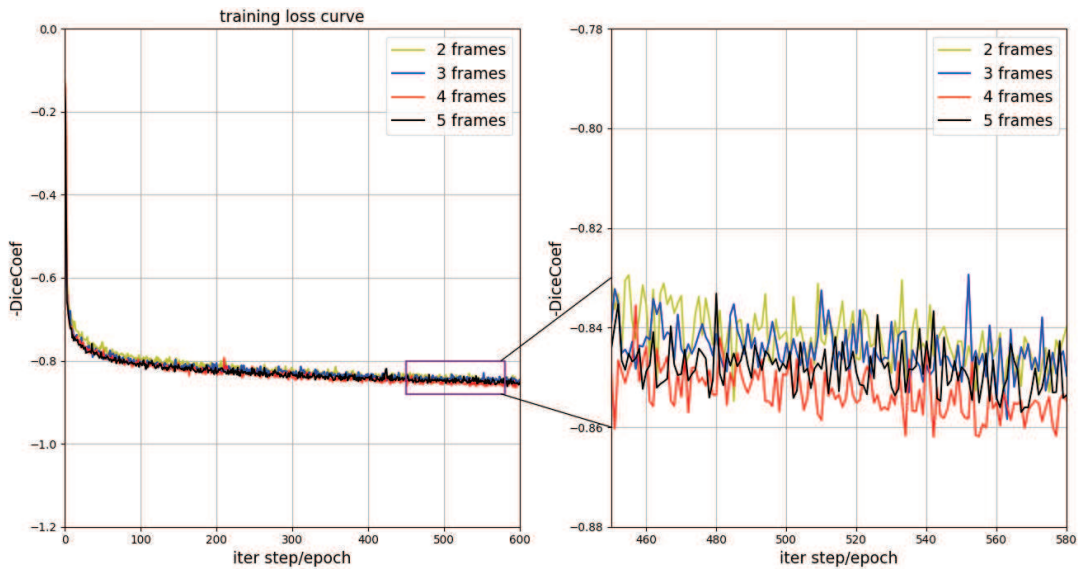}
 	\caption{The training loss curve (left) and its local enlarged curve (right) for different input strategies in training process. The 4 frame's input strategy can achieve the least training loss.}
 	\label{Fig2}
 \end{figure*}

To verify the rationality of the input configuration of 4 frames, we respectively input successive frames, \textit{i.e.}, 2, 3, 4, 5 frames, into the network to investigate the network convergence performance. As shown in Fig. \ref{Fig2}, there are slightly differences in terms of the convergence results of loss function (DiceCoef) in the training set. The smaller that the loss becomes, the better the fitting performance that the model achieves. When we input 4 frames into the network, the loss converges at about -0.86, which is the smallest compared with other input strategies. Therefore, 4 frames are reasonable and feasible to the input configuration for accurate segmentation results. 

\begin{figure*}
	\centering
	\includegraphics[width=1.2\linewidth]{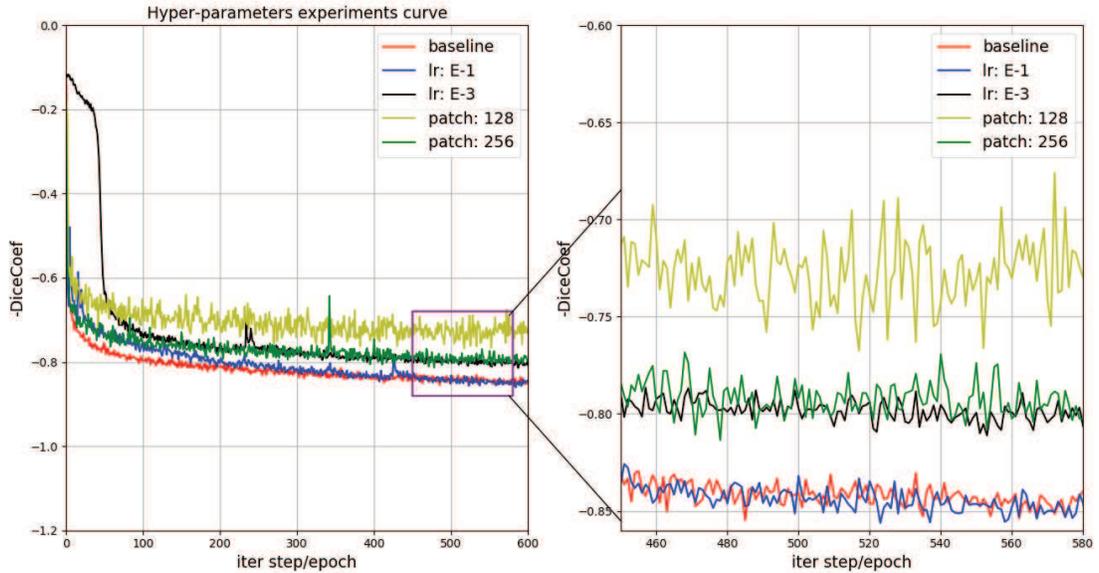}
	\caption{The hyper-parameter experiments (left) and its local magnified curve (right) comparing different learning rates (lr) and input sizes with baseline (lr: 0.01, input size: $512\times512$.)}
	\label{Fig3}
\end{figure*}

We further explore hyper-parameter setup in a sensitivity analysis of the parameters including the learning rate and the size of input images. In our baseline model, we experientially set up learning rate and input size as 0.01 and $512\times512$ respectively. In the subsequent contrast experiments, we merely change either learning rate (\textit{i.e.}, 0.1, 0.001) or input size (\textit{i.e.}, $128\times128$, $256\times256$) and make other experimental configurations remain unchanged. As shown in Fig. \ref{Fig3}, the learning rate and input size make a big difference to the training loss. The smaller input size means the limited receptive field so that the model suffers from performance degradation. Furthermore, the smaller initial learning rate may cause the model to get stuck into local minima in optimization, which also decreases the segmentation performance. By these experiments, we can conclude that our settings of learning rate (0.01) and the input size ($512\times512$) lead to the best performance considering both speed and accuracy.

\subsection{Modified U-net architecture}
We employ U-net as the fundamental architecture in SVS-net. U-net \citep{Ronneberger15} is a classical and powerful segmentation network architecture widely used for biomedical images by effectively exploring the underlying high-resolution and low-resolution information in biomedical image. Using skip-layers to build a bridge transmitting multi-scale information from encoder network to decoder network, U-net can improve the spatial accuracy of a deep CNN for final high-resolution segmentation results. 

On the one hand, XCA images have low contrast and fuzzy boundary, which require more high-resolution detail information for accurate segmentation. The skip connection mechanism in U-net allows high-resolution information delivery to the decoder network for detail recovery. On the other hand, the internal tissue structures with their topologies in XCA images are relatively fixed, the distribution of segmentation targets in the XCA images is regularly presented with simple and clear semantics, which require more low-resolution information to present accurate semantic information for the target object recognition. Multiple downsampling operations in U-net's encoder network appropriately provide low-resolution information for contextually semantic recognition. In Fig. \ref{Fig1}, the encoder network capture 3D temporal-spatial contexts through 3D convolutions followed by 3D residual convolutional blocks except the last convolutional operation. The decoder network enables precise localization of high-resolution target vessel semantic information via upsampling layers and CAB. 

Aiming at the accurate 2D+t XCA vessel segmentation, we make following adaptations based on conventional U-net: (1) In the encoder network, there are 7 stages of 3D convolution. The first six convolutional stages followed by 3D residual convolutional block (in Section 2.4) are utilized to extract rich temporal-spatial feature representations, which provide contexts for subsequent vessel mask inference in the decoder network. The output of each 3D residual convolutional block is passed to the next 3D convolutional stage and the skip connection layer respectively. At the last two 3D convolutional stages, spatial dropout (0.5) is employed before executing convolution to avoid overfitting; (2) In the skip connection layers, we fuse temporal-spatial feature representation by mapping from 3D space to 2D space via $4\times 1\times 1$ convolutional kernel in FFO, where the first dimension of convolutional kernel indicates the temporal axis, \textit{i.e.}, 4 indicating 4 channels (frames) in the temporal domain. The temporal domain features are then fused together by temporal axis convolution. The FFO can be formulated as follows:
\begin{equation}\label{skip}
X_{F_{l}}=Squeeze(X_{F} \otimes W)
\end{equation}
where $ X_{F} \in \mathbb{R}^{C\times T \times H \times W}$ is the spatial-temporal feature map coming from the output of each 3D convolutional stage in the encoder network, $ X_{F_{L}} \in  \mathbb{R}^{C\times H \times W}$ denotes fusing spatial-temporal feature map, $C,T,H,W$ are the features' channel dimension, temporal dimension, height, and width, respectively. $W$ denotes $4\times 1\times 1$ convolutional kernel, $\otimes$ represents convolution operation, Squeeze denotes dimension compress, a straightforward schematic can be seen in the left bottom of Fig. \ref{Fig1}; (3) In the decoder network, to gradually recover the feature maps’ spatial resolution, we take advantage of the parameter-free bilinear upsampling strategy rather than transposed convolutional operations, which contributes to reduce the number of trainable parameters without degrading the segmentation performance \citep{de2018clinically}. Each upsampling layer is followed by one CAB (see the right bottom of  Fig. \ref{Fig1}) and one 2D residual convolutional block (Block2D, see Fig. \ref{Fig1}). Note that the high-stage and low-stage feature map outputs with the same resolution from the upsampling layer and the skip connection layer are inputted simultaneously to CAB (as illustrated in the right bottom of Fig. \ref{Fig1}), which is employed to learn the most discriminative features from noisy and complex background artifacts (see the detail in Section 2.5). After the last 2D residual convolutional block, we employ $1\times1$ convolution followed by sigmoid activation function to yield the final vessel mask.

\subsection{2D and 3D Residual convolutional blocks}
Generally speaking, increasing the depth of networks can improve network generalization capacity. However, a very deep network implies the difficulty in promoting gradient back propagation, which results in the poor performance. To overcome this problem, He \textit{et al.} \citep{he2016deep} develop the deep residual network to facilitate gradient back propagation by identity mapping connection. Zagoruyko \textit{et al.} \citep{zagoruyko2016wide} demonstrate that the two stacked convolutional layers in single residual block is optimal architecture compared with other settings. Hence, we follow the strategy as advised in \citep{zagoruyko2016wide} and employ 3D residual blocks and 2D residual blocks in encoder and decoder networks respectively. 

\subsection{Channel attention mechanism}
\begin{figure*}
	\centering
	\includegraphics[width=1.2\linewidth]{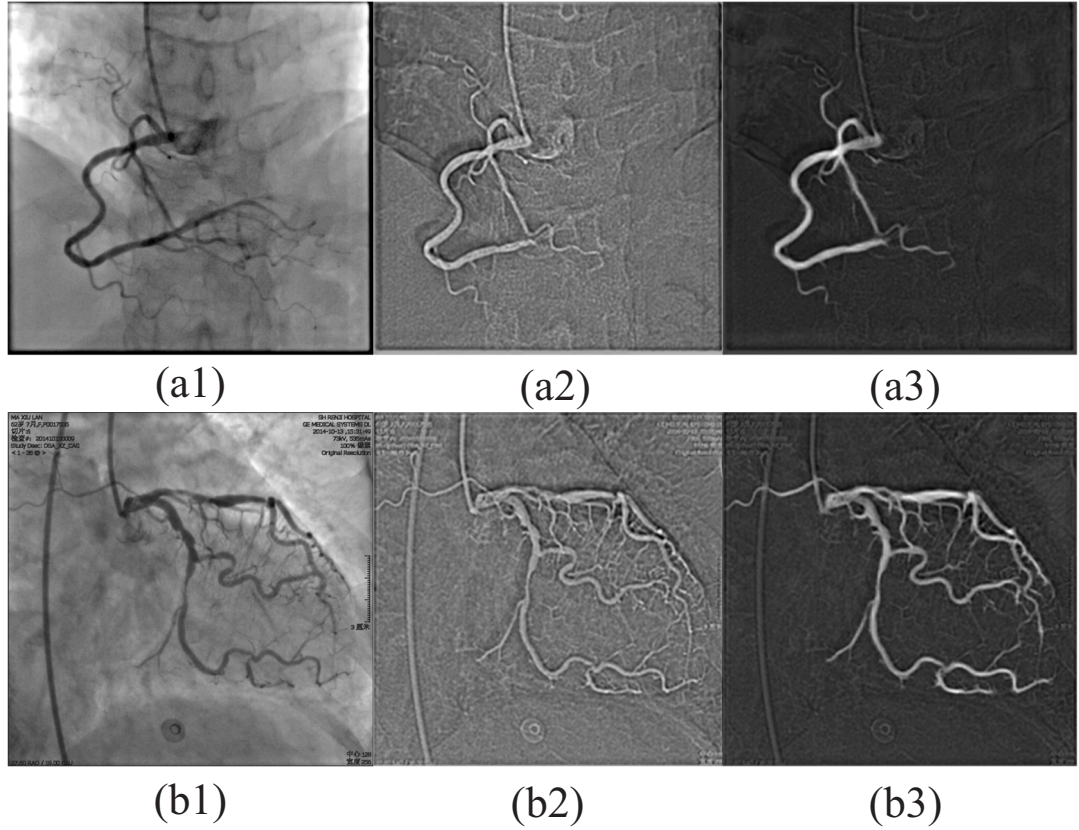}
	\caption{Two instances of feature visualizations for illustrating the CAB's effects: suppresses the noises in the background areas while highlights the foreground vessel feature. From left to right, each row displays the original XCA image; the 2nd channel of fused spatial feature maps in the output of the second skip connection layer (Fig. \ref{Fig1}) before inputting to the CAB, it contains noise pollution from the background areas; the 2nd channel of refined feature maps from the output of CAB in the decoder stage (Fig. \ref{Fig1}). The background noise is reduced and the foreground vessel feature is highlighted via the channel attention operation.}
	\label{Fig4}
\end{figure*}

To learn more rich and multi-scale feature representation for extracting vessels from complex and dynamic background artifacts, the proposed SVS-net firstly extracts multiple types of features by multiple convolutional kernels in every convolutional stage of the encoder stage (see Fig. \ref{Fig1}). Note that there exists three problems: (1) each channel of feature maps represents one specific feature type but not all features are equally significant to the final output; (2) XCA sequence contains not only the target vessels but also much disturbance of overlapping structures that have similar appearances and intensities to vessels, these disturbances are aggregated nearly at different positions with their relatively various moving speeds. Therefore, these disturbances are distributed in different feature channels; (3) Note that the skip connection layers fuse the temporal-spatial features through $4\times 1\times 1$ convolution, which performs linear combination in temporal domain. This combination inevitably introduces extra noisy artifacts into different feature channels besides the noisy disturbances inherent in the XCA sequences. As shown in Fig. \ref{Fig4}(a2)(b2), the fused spatial feature map contains a lot of noise artifacts from the background area, which may decrease the accuracy of vessel detection. Therefore, the fused spatial feature map from the output of skip connection layer should be well treated to weaken the noise disturbance from the noisy backgrounds and emphasize the vessel feature simultaneously. To this end, we introduce an effective scheme called as channel attention mechanism for highlighting foreground vessel features and noise removal. 

Through the operation of CAB, the SVS-net can adaptively highlight some channel information meanwhile suppress the trivial channel information. Hence, the predicted probability map is gradually improved. Inspired by the works \citep{hu2018squeeze,yu2018learning}, we introduce the CAB to weight the feature maps from the low-stage output from the skip connection layer and then combine with the corresponding high-stage feature maps that are outputted from the upsampling layer. High-stage output feature maps contain more advanced global semantic information while low-stage feature maps contain more detailed yet noisy information, therefore the high-stage features can provide clues to screen useful information from low-stage feature maps and generate more pure feature representation. Under the guidance of high-stage features, the attention weights are learned and used to obtain discriminative salient features. As shown in Fig. \ref{Fig4}(a3)(b3), the low-stage feature map from the output of skip connection layer is refined by the CAB. From the Fig. \ref{Fig4}(a2)-(a3) and Fig. \ref{Fig4}(b2)-(b3), the background noises in Fig. \ref{Fig4}(a2)(b2) are greatly reduced while the foreground vessel features are highlighted in Fig. \ref{Fig4}(a3)(b3).

Specifically, the CAB do the following operations (see the right bottom of Fig. \ref{Fig1}): the low-stage feature maps $ X_{F_{l}} \in \mathbb{R}^{C \times H \times W}$ and the corresponding high-stage feature maps $ X_{F_{h}} \in \mathbb{R}^{C \times H \times W} $ are concatenated together to make feature maps $ X_{F} \in \mathbb{R}^{2C \times H \times W}$. Furthermore, a global average pooling is performed on the concatenated feature maps to generate the weighted vector $ W_{X_{F}} \in \mathbb{R}^{2C \times 1 \times 1}$.\citep{yu2018learning}. Two $1\times1$ convolutional operations, which are followed by the rectified linear unit function and sigmoid function, respectively, are performed on  $ W_{X_{F}} \in \mathbb{R}^{2C \times 1 \times 1}$ to learn inter-channel relationship and the final channel attention weights vector $ W_{X_{F_{l}}} \in \mathbb{R}^{C \times1\times1}$ is achieved. The obtained attention vector multiply low-stage feature maps in channel-wise manner, then the weighted feature maps from low stage are added with the corresponding high-stage feature maps to be subsequently passed to the next layer. The whole process of generating attention weights can be expressed as:
\begin{equation}\label{attention}
W_{X_{F_{l}}}=\phi(\varphi(GAP( X_{{F}}))) 
\end{equation}
where $GAP$ means the operation of global average pooling, $\varphi$ denotes $1\times1$ convolution followed by rectified linear unit and $\phi$ indicates $1\times1$ convolution followed by sigmoid activation. An intuitive display of CAB is shown in the left bottom of Fig. \ref{Fig1}.

\subsection{Data augmentation}
As there are limited manually annotated datasets, data augmentation is necessary for the benefit of improving the model generalization. To teach SVS-net how to accommodate to various sample transformations, we adopt multiple augmentation methods including rotations by the angle ranging in $[-10^\circ,10^\circ]$, flipping both horizontally and vertically, scaling by a factor of $0.2$, random crop, affine transformations. For the images in our dataset, there is a 50\% probability to perform each of above transformations to generate new samples in real time during the training process.  

\subsection{Loss function}
To tackle the class imbalance problem in vessel segmentation, we employed Dice loss function to guide parameters learning. The class imbalance problem mainly has two aspects: firstly, the number of negative pixels (being 0, \textit{i.e.}, background) is much more than the number of positive pixels (being 1, \textit{i.e.}, vessel pixels); secondly, the ratio between the two classes varies a lot among both inter-frame in the same XCA sequence and intra-frame or inter-frame in different XCA sequences. Currently most semantic segmentation tasks adopted the following cross entropy (CE) \citep{ronneberger2015u,8578429} to optimize the network: 
\begin{equation}\label{LOSS1}
L_{CE} = \sum_{1}^N y_{i}\log p_{i}+(1-y_{i})\log(1-p_{i})
\end{equation}
It can be observed that from Equation (3), each pixel contributes equally to the CE loss. Hence, CE loss tends to bias the network's optimization.

Different from CE loss calculated in pixel-wise form, Dice loss can avoid above problem by measuring the overlap ratio between ground truth mask and the predicted vessel mask. Dice loss is defined in \citep{drozdzal2018learning,8328863} as follows:
\begin{equation}\label{LOSS2}
L_{DiceCoef} =- \frac{2\sum_{1}^N p_{i}y_{i}+ \epsilon}{\sum_{1}^N p_{i}+\sum_{1}^N y_{i}+\epsilon}
\end{equation}
where $y_{i} \in \{0,1\}$ is ground truth label and $p_{i} \in [0,1]$ is predicted value for location $ i $. $ N $ is the total number of pixels, $ \epsilon $ is a very small constant used to keep value stable. From Equation (4) we can find that the Dice loss is applied to the whole mask and it measures the overall loss for that mask rather than the average loss across all the pixels.

\section{Experiment results}
\subsection {Materials}
In our experiments, 120 sequences of real clinical X-ray coronary angiograms images are acquired from Ren Ji Hospital of Shanghai Jiao Tong University. The length of each sequence ranges from 30 to 140 frames. Images from 120 sequences have been manually annotated by three experts to constitute the ground truth. Specifically, for the totally hard-annotated 323 samples from these 120 annotated sequences including extremely low-contrast vessels and vessel trees that contain a lot of thin vessel branches, we take three experts' average annotated result as the final ground truth.

It is worth noting that these XCA sequences in the dataset are acquired from different machines (\textit{i.e.}, 800 mAh digital silhouette angiography X-ray machine from Siemens, medical angiography X-ray system from Philips), the resolution, the noise distribution and the pixels' intensity range of each single frame are different. To eliminate these differences, we resize the images from the XCA sequence into $512 \times 512$ resolution with 8 bits per pixel, employ Poisson denoising methods \citep{cerciello2012comparison} to smooth the noise and normalize the pixels' intensity range into $0-1$. 

Furthermore, due to the varieties of XCA images with different directions and angles of X-ray penetration as well as different patient sources with different dosages of contrast agents, the vessels visibility of different sequences in clinic is quite changeable. Thus, designing a robust vessel segmentation algorithm is necessary for the XCA data with poor visual quality. Besides, proper selection of frames from each sequence for experiment is crucial \citep{lim2018learning} especially when both of the background and foreground are dynamic and contain many artifacts. The strategy of selecting the training frames is similar to \citep{wang2017interactive}. We selected XCA images containing most of vessel structures as experiment samples from 120 annotated sequence according to their lengths and  visual quality. Totally, 332 samples are obtained for our experiment. The dataset is randomly divided into training dataset, validation dataset, and test data at approximately 0.5, 0.25 and 0.25, respectively. 

\begin{figure*}
	\centering
	\includegraphics[width=1.2\linewidth]{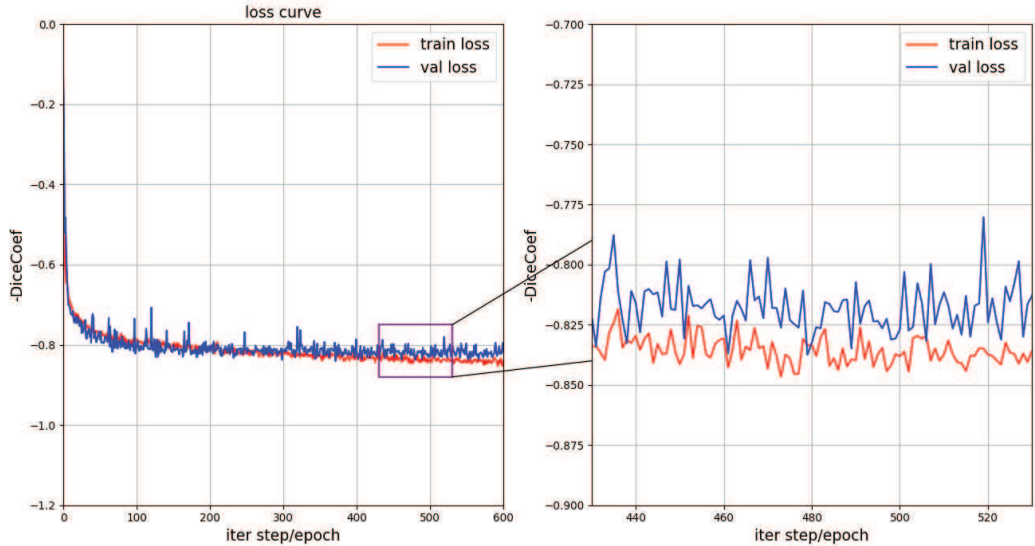}
	\caption{The loss curve and its local enlarged curve for both training set and validation set in training process}
	\label{Fig5}
\end{figure*}

We investigate the proposed model's performance on the abovementioned dataset. We plot the loss curves for both the training set and validation set in the training process. As can be seen in Fig. \ref{Fig5}, for both training set and validation set, the loss reduces quickly at the beginning stage of training process, and gradually converges. There is no sign that the model falls into over-fitting or under-fitting state. Meanwhile, the size of our dataset is assumed to be properly matched into the size of our model. 

All the experiments performed in this work were approved by our institutional review board. The dataset which will be released to public has received the transfer agreement from our cooperative partners.  All the dataset is stored in mat array format according to the corresponding filenames, and they will be available on website\footnote{The source codes and dataset will be available at \url{https://github.com/Binjie-Qin/SVS-net}}. You can also visit the website to access further detailed information on the dataset. 

\subsection {Evaluation metrics}
Several metrics, namely, detection rate (DR), precision (P), and F measure are employed to quantitatively evaluate the performances of our segmentation method and also compare them with other state-of-the-art methods. The above metrics are defined as below:
\begin{equation}\label{DRPF}
DR = \frac{TP}{TP+FN},\ P = \frac{TP}{TP+FP},\ F = \frac{2*DR*P}{DR+P}
\end{equation}
where TP (true positives) is the total number of correctly classified pixels in vessel regions of the predicted vessel probability map, FP (false positives) indicates the total number of wrongly identified as vessel pixels but practically belonging to background in the predicted vessel probability map, TN (true negatives) and FN (false negatives) represent the total number of correctly predicted as background pixels and wrongly predicted as background pixels in the predicted output, respectively. DR measures the proportion between the correctly identified vessel pixels and the total vessel pixels in the ground truth, P measures the ratio of true positives among all the true positives. F measure comprehensively considers both P and DR metrics and indicates the overall segmentation performance. All these metrics range in $\left[ 0,1 \right]$, and a higher value indicates better segmentation performance. 

\subsection{2D vs 3D with and without channel attention mechanism} 
We utilize 3D convolution layers to extract rich temporal-spatial feature representation in encoder network. To investigate whether the temporal-spatial features are more advanced compared to purely spatial features extracted by 2D convolutional layers for generating final predicted probability map, we replace 3D convolutional layers with corresponding 2D ones in encoder network while keeping the decoder network the same. It is noted that simple substitution reduce the number of trainable parameters and hence weaken the model's expressive capacity. For fair comparison, we increase the number of convolution layers in the encoder network for 2D version to make both 3D version and 2D version have comparable amount of parameters. In addition, we investigate the effectiveness of CAB by removing it from decoder network in SVS-net and 2D model respectively. We choose our 2D model without CAB (2D naive) as baseline, and compare it with 2D model with CAB (2D+CAB), 3D model without CAB (3D naive) and 3D model with CAB (3D+CAB) respectively. 
\begin{table}
	\caption{The average detection rate (DR), precision (P) and F measure (mean value $ \pm $ standard deviation) for test data.}
	\label{DRPF}       
	\begin{tabular}{ l | p{1.4cm} p{1.4cm} p{1.4cm} }
		\hline\noalign{\smallskip}
		Method & DR & P & F   \\
		\noalign{\smallskip}\hline\noalign{\smallskip}
		2D naive & 0.7640 $\pm$0.0701 & 0.8615 $\pm$0.0694 & 0.8056 $\pm$0.0431 \\
		2D+CAB & 0.7638 $\pm$0.0738 & 0.8595 $\pm$0.0684 & 0.8046 $\pm$0.0459 \\
		3D naive & 0.7959 $\pm$0.0714 & 0.8640 $\pm$0.0586 & 0.8255 $\pm$0.0714 \\
		3D+CAB & \textbf{0.8424 $\pm$0.0813} & 0.8492 $\pm$0.0605 & \textbf{0.8428 $\pm$0.0531} \\
		\noalign{\smallskip}\hline
	\end{tabular}
\end{table}
\begin{figure}
	\centering
	\includegraphics[width=0.9\linewidth]{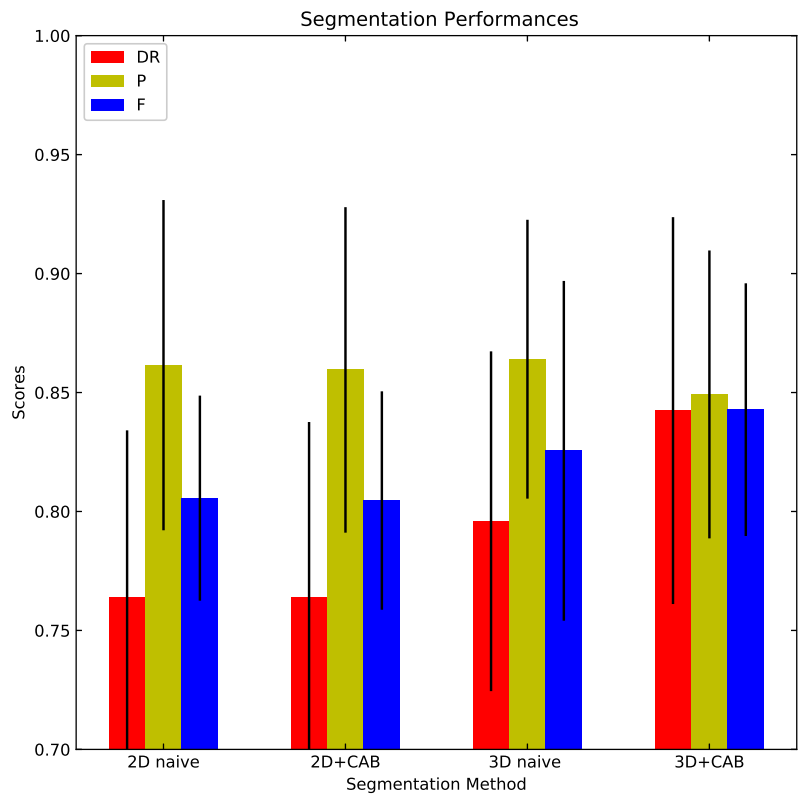}
	\caption{Vessel segmentation performance using 2D and 3D convolutional layers with and without CAB. The detection rate (DR), precision (P) and F measure of test data. }
	\label{Fig6}
\end{figure}

To quantitatively evaluate the performance, we measure three metrics on the test set. The results are shown in Table \ref{DRPF} and Fig. \ref{Fig6}. Specifically, we compare the performance between different feature extraction manners and then analyze the function of CAB. As can be seen in Table \ref{DRPF}, compared with the baseline, the 3D naive model obtains higher scores in terms of DR (79.59\%), P (86.40\%), F (82.55\%) measures and surpasses its 2D counterpart by 4.17\%, 0.29\% and 2.47\%, respectively. Despite they have almost similar model complexity, their performance have big differences. This is because 3D version integrating contextual spatial-temporal features while 2D version only using spatial domain features to predict the final probability map. Obviously, the former one has more sufficient and robust information in discriminating between noisy artifacts and vessel trees. 

Next, we investigate the effect of CAB in both 3D and 2D scenarios. From Table \ref{DRPF}, we can find that the channel attention strategy decreases the P measures from 0.8615 to 0.8595 and reduces the DR scores from 0.7640 to 0.7838 in 2D case. There are slight changes in both DR and P. Hence, F measures almost keep consistent due to F measure achieving the trade-off between P (related with FPs) and DR (related with FNs) measures. In 3D case, the CAB also shows the good compromise between P and DR (\textit{i.e.}, DR increases from 0.7959 to 0.8424 while P declines from 0.8640 to 0.8492) but the overall performance F measure improves by 2.09\% and arrives at 0.8428. We analyze the role that the CAB plays in the overall performance: in the 2D case, we find the fact that CAB makes a compromise between P and DR with rare improvement in F measure, which is now interpreted that 2D version failing to provide sufficient and valuable information for CAB to choose from; however, in 3D case, the temporal-spatial information is relatively rich so that the CAB can suppress trivial features in noisy background and utilize the most discriminative ones in foreground to generate fine vessel mask. 

Furthermore, we analyze the stability of SVS-net's segmentation performance. When we compare SVS-net (3D+CAB) with the baseline (2D naive), the DR and F witness a large increase by 10.26\% and 4.61\%, respectively. The metrics have a relatively high standard deviation, which indicates that there exist both relatively hard-segmented and relatively easy-segmented samples in test sets. The relatively hard-segmented samples may pull down the whole metrics and result in the high standard deviation. In the future, we will enlarge our dataset and increase the number of hard-segmented samples to help SVS-net pay more attention to them for improving the its performance on those samples. It is worth noting that the relatively high standard deviation problem almost exists in all the tested methods, which illustrate there indeed exists hard-segmented samples. The comparative methods also fail to deal with these hard-segmented samples. In view of the average metrics, we further our confidence that SVS-net performs better than all the other methods.

Moreover, the 3D+CAB model achieves the highest DR score while the lower P score. Higher DR score means lower FNs, which indicates SVS-net have superior capacity in detecting vessel pixels from backgrounds. In detecting vessel pixels, it is likely to mistakenly identify backgrounds that resemble vessel pixels as vessel pixels. This may increase FPs to some degree, so that the P score will suffer from degradation. However, the 3D+CAB model's overall F measure performance is highest, which illustrates its balanced and better performance in accurately recognizing both vessel pixels and background pixels when compared with other methods. 
\begin{figure*}
	\centering
	\includegraphics[width=1.1\linewidth]{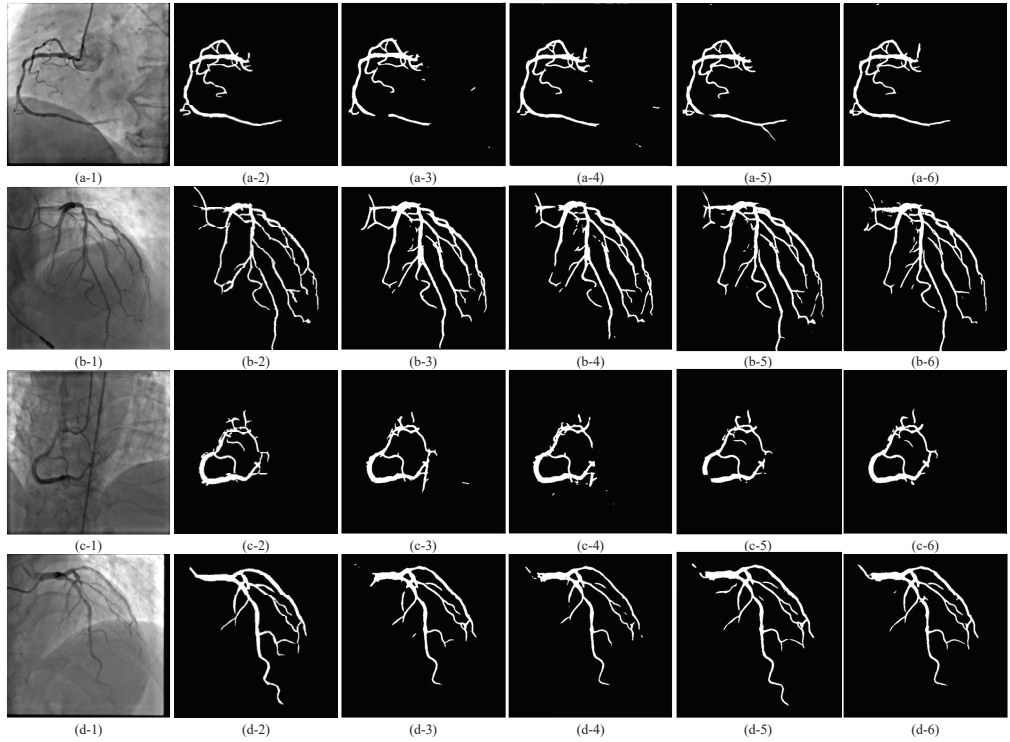}
	\caption{Four instances of vessel segmentation result by different vessel segmentation methods. From left to right, each row displays the original XCA image, the manually outlined ground truth vessel segmentation, the vessel images segmented by 2D naive, 2D+CAB, 3D naive, 3D+CAB, respectively.}
	\label{Fig7}
\end{figure*}

Intuitive quantization result can be seen in Fig. \ref{Fig6}. The typical segmentation results are displayed in Fig. \ref{Fig7}. As shown in Fig. \ref{Fig7}, the vessel masks produced by SVS-net have less fractures (\textit{i.e.}, less FNs) than do the other methods, which implies SVS-net's better performance on detecting vessel pixel (TPs). In 2D settings, the vessel masks are either have more fracture (\textit{i.e.}, more FNs) or have more artifacts (\textit{i.e.}, more FPs). This phenomenon shows that their pool ability on differentiating vessel pixels from background pixels. Through above observation and analysis, we validate the effectiveness of the 3D convolutional layers and CAB used in SVS-net.
\begin{table}
	\caption{The average detection rate (DR), precision (P) and F measure (mean value $ \pm $ standard deviation) for test data.}
	\label{loss}       
	\begin{tabular}{ l | p{1.4cm} p{1.4cm} p{1.4cm}  }
		\hline\noalign{\smallskip}
		Method & DR & P & F   \\
		\noalign{\smallskip}\hline\noalign{\smallskip}
		CE loss & 0.8197 $\pm$0.0814 & 0.8423 $\pm$0.0633 & 0.8262 $\pm$0.0428 \\
		Dice loss & \textbf{0.8424 $\pm$0.0813} & 0.8492 $\pm$0.0605 & \textbf{0.8428 $\pm$0.0531} \\
		\noalign{\smallskip}\hline
	\end{tabular}
\end{table}

\subsection{Cross entropy loss vs Dice loss}
To validate the loss function in deep network for vessel segmentation, we employ CE loss and Dice loss function as objective function to train our model respectively and use the same settings with other parts of network. It can be seen in Table \ref{loss}, the model trained with Dice loss can result in the performance improvement by 2.0\% in terms of F measure, which implies the effectiveness of Dice loss in class imbalance segmentation task when compared with the CE loss. Besides, the model trained with Dice loss achieves higher F measure and DR score but slightly higher P score when compared with the model trained with CE loss. The Dice loss's optimization goal is expected to facilitate the model in getting higher gain of F measure for achieving better overall segmentation performance. When the model is optimized, the overall segmentation performance of F measure certainly has an upper bound. With this upper bound on the F measure that is computed from the harmonic mean of DR and P, the model makes a trade-off balance between the DR and P metrics, so that it may results in the balance with higher DR and slightly higher P.
\begin{figure*}
	\centering
	\includegraphics[width=1.0\linewidth]{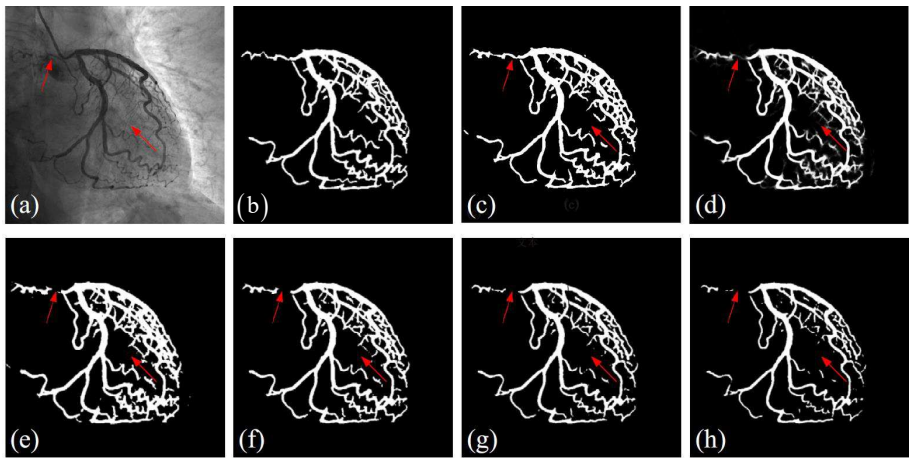}
	\caption{Original vessel segmentation result by Dice loss, CE loss and threshold (0.2, 0.4, 0.6, 0.8 respectively) postprocessing for the segmentation result by CE loss. (a) Original XCA image. (b) Manually outlined ground truth. (c) Original segmentation result by Dice loss. (d) Original segmentation result by CE loss. (e) 0.2 thresholding result of (d). (f) 0.4 thresholding result of (d). (g) 0.6 thresholding result of (d). (h) 0.8 thresholding result of (d)}
	\label{Fig8}
\end{figure*}

Then, we compare the intuitive segmentation results by two loss functions. When compared with the vessel masks obtained from the model trained with Dice loss, the vessel masks obtained from the CE loss have blurred boundaries (see the red arrows in Fig. \ref{Fig8}(c)-(d)), which means pixels at vessel boundaries are less confident to discriminate whether they belong to vessels or backgrounds. Therefore, to get binary vessel masks, we should carefully apply proper threshold to the original probability maps which involves troublesome manual operations (shown in Fig. \ref{Fig8}(e)-(h))). While the masks produced by model trained with Dice loss have clear boundaries, there is no need to utilize threshold anymore. Therefore, the Dice loss function is appropriate to train SVS-net for sequential vessel segmentation.

\subsection{Comparison with other state-of-the-art methods}
We compare SVS-net with three traditional vessel segmentation algorithms, \textit{i.e.}, Coye’s filter method (Coye's)\footnote{\url{http://www.mathworks.com/matlabcentral/fileexchange/50839}} \citep{coye2017novel}, Jin’s spatially adaptively filtering method (Jin's) \citep{jin2018low}, Kerkeni’s multi-scale region growing method (Kerkeni's) \citep{KERKENI201649}, and four deep learning-based methods, \textit{i.e.}, Retinal-net\footnote{\url{https://github.com/orobix/retina-unet}} \citep{liskowski2016segmenting,ronneberger2015u}, bridge-style U-Net with salient mechanism (S-UNet)\footnote{\url{https://github.com/hdd0411}}\citep{Hu2019},  X-ray net\footnote{\url{https://github.com/pambros/CNN-2D-X-Ray-Catheter-Detection}} \citep{ambrosini2017fully}, short connected deep supervised net (BTS-DSN)\footnote{\url{https://github.com/guomugong/BTS-DSN}} \citep{guo2018deeply}. 
\begin{table}
	\caption{The average detection rate (DR), precision (P) and F measure (mean value $ \pm $ standard deviation) for test data by state-of-the-art methods and our method.}
	\label{Time}       
	\begin{tabular}{ l | p{1.4cm} p{1.4cm} p{1.4cm} p{1.4cm} p{1.4cm} }
		\hline\noalign{\smallskip}
		Method & DR & P & F &param & Inference Time  \\
		\noalign{\smallskip}\hline\noalign{\smallskip}
		Coye's & 0.5694 $\pm$0.3096 & 0.2127 $\pm$0.1365 & 0.2963 $\pm$0.1752 & \  &0.071s  \\
		Jin's & 0.6127 $\pm$0.1948 & 0.7715 $\pm$0.2126 & 0.6639 $\pm$0.1677  & \ &11.61s \\
		Kerkeni's & 0.6703 $\pm$0.1322 & 0.7348 $\pm$0.1321 & 0.6863 $\pm$0.1047 &\ &4.708s \\
	    Retinal-net & 0.7708 $\pm$0.1003 & 0.6807 $\pm$0.1160 & 0.7141 $\pm$0.0865  &0.47M &2.28s\\
	    SU-Net & 0.6914 $\pm$0.1057 & 0.9018 $\pm$0.0785 & 0.7734 $\pm$0.0580 &15.32M &0.046s \\
	    X-ray net & 0.7974 $\pm$0.0748 & 0.7780 $\pm$0.1038 & 0.7794 $\pm$0.0586  &14.1M &0.053s\\
		BTS-DSN & 0.7251 $\pm$0.0971 & 0.8626 $\pm$0.0750 & 0.7803 $\pm$0.0612 &7.8M &0.067s \\
		SVS-net & \textbf{0.8424 $\pm$0.0813} & 0.8492 $\pm$0.0605 & \textbf{0.8428 $\pm$0.0531} &10.2M &0.178s \\
		\noalign{\smallskip}\hline
	\end{tabular}
\end{table}
\begin{figure*}
	\centering
	\includegraphics[width=0.8\linewidth]{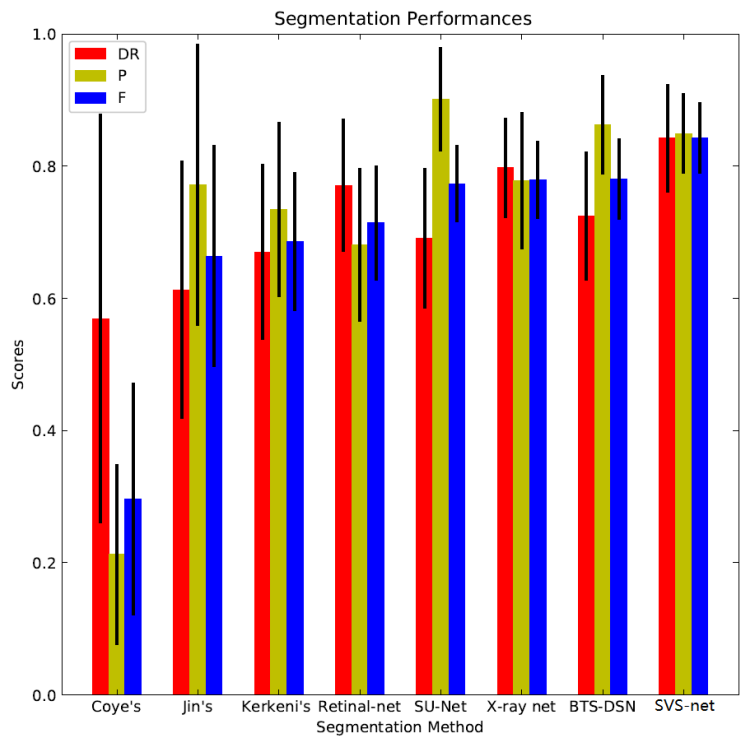}
	\caption{Vessel segmentation performance using different methods. The detection rate (DR), precision (P) and F measure of test data. }
	\label{Fig9}
\end{figure*}
\begin{figure*}
	\centering
	\includegraphics[width=1.2\linewidth]{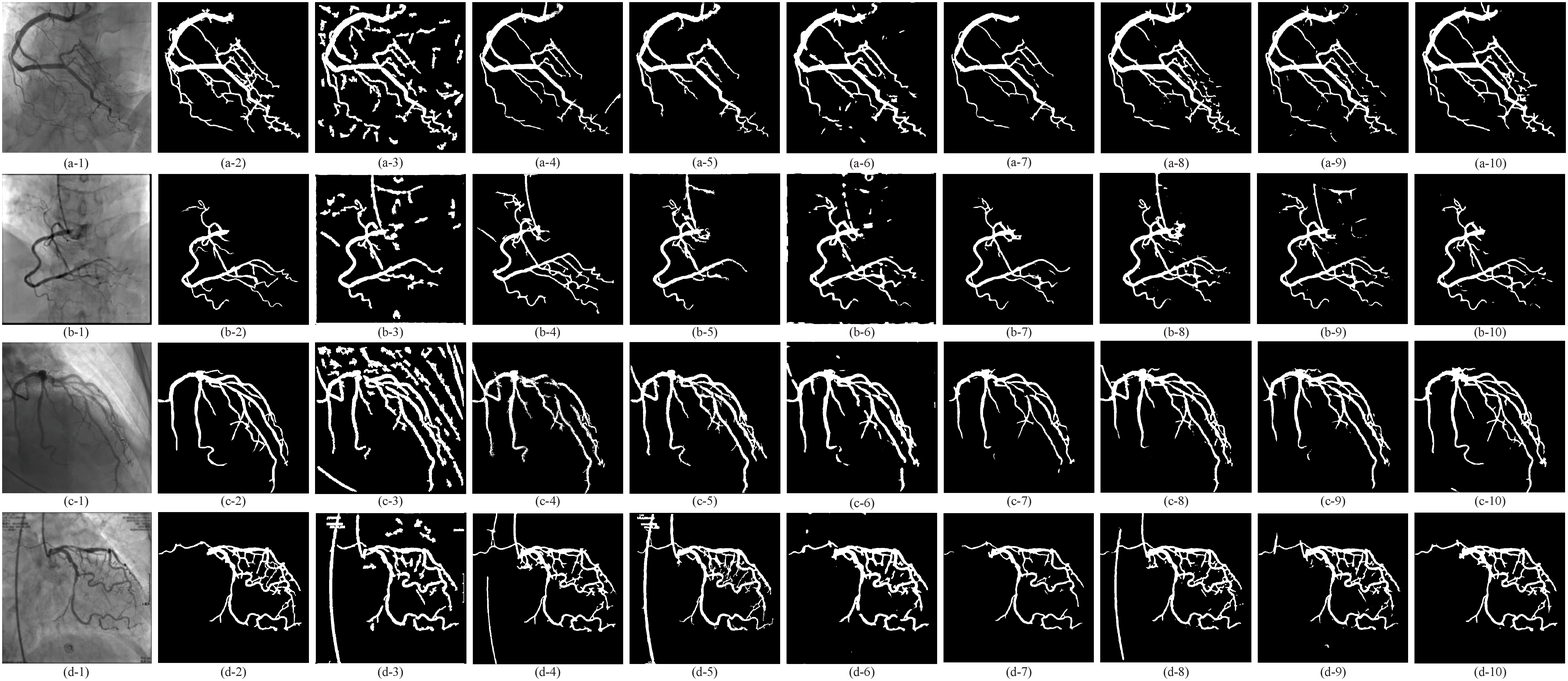}
	\caption{Four instances of vessel segmentation for real XCA image sequence by different vessel segmentation methods. From left to right, each row displays the original XCA image, the manually outlined ground truth vessel segmentation, the vessel images segmented by Coye's, Jin's, Kerkeni's, Retinal-net, SU-Net, X-ray net, BTS-DSN, and SVS-net, respectively.}
	\label{Fig10}
\end{figure*}

Table \ref{Time} and Fig. \ref{Fig9} summarize the segmentation performances for the different approaches. It can be observed that SVS-net surpasses other methods by a large margin on all the metrics. The segmentation result comparison is given in Fig. \ref{Fig10}. Among traditional vessel segmentation methods, Jin's and Kerkeni's obtain relatively better quantitative metrics and visual qualities. However, Jin's algorithm detects more FPs due to its inaccurate local and global thresholds applied to the low contrast XCA images. Kerkeni's method can well segment thick vessels out but almost fail to recognize thin ones. This is because thin vessels are extremely indistinct compared with thick vessels and region growing algorithm cannot assign seed points to these regions so that they wrongly exclude these thin vessels from vessel regions in the subsequent procedure. 

In regard to deep learning methods, they surpass traditional methods and gain higher performances in term of some metrics. Retinal-net, as a patch based method, introduces more background residuals since it lacks more global contextual information to guide the segmentation. X-ray net inputs current frame image with its first three frames to the network, but it simply concatenates them together and can not effectively extract temporally consistent information. It not only increases temporal information but also introduces disturbances at the same time. BTS-DSN adopts deeply supervised strategy and achieves relative higher metrics. However, there are still FPs in the vessel region.  Compared with above deep network methods, SVS-net can not only robustly detect the vessel regions with almost intact vessel structures with continuous vessel branches but also effectively remove the noisy background artifacts. The continuity and integrity of the segmented vessel branches is assumed to be owed to the contextual information inferred in the temporal-spatial features extracted by the encoder network and feature fusion in the skip connection layers. The noise reduction in the segmented vessel regions is mostly derived from the discriminative feature selection implemented by the channel attention mechanism. Therefore, the temporal-spatial feature extraction, feature fusion and the discriminative feature learning adopted in SVS-net are necessary to help improve the segmentation performances.
	
Additionally, there is a small number of thin vessel branches fail to be recognized by SVS-net. It is really challenging and we plan to design novel loss function in our future work, which will differentiate the thick and thin vessels efficiently and integrate these different vessels with different weights. In this way, we increase the weights of thin vessels in loss function and promote the model to pay more attention to the thin vessels. Thin-vessel segmentation is definitely a promising direction for improving the clinical value of XCA images.

Our experiments are implemented on GPU (\textit{i.e.}, NVIDIA 1080 Ti, 11GB). The number of parameters and the average runtime of per test image for deep-learning methods are listed in Table \ref{Time}. Compared with other deep learning-based methods having bigger or fewer number of parameters, \textit{i.e.}, from the 15.32 million parameters for SU-Net to the 0.47 million parameters for Retinal-net, as well as having longer or shorter inference time, \textit{i.e.}, from the 2.28 seconds for Retinal-net to the 0.046 seconds for SU-Net, SVS-net has 10.2 million parameters and 0.178 second inference times to achieve an intermediate level of complexity. The reason for SVS-net's medium-complexity in achieving its best segmentation performance is two-fold: 1) the 3D convolutional layers instead of 2D convolutional layers are adopted in the stage of feature extraction; 2) the fully connection layers are utilized in the stage of feature refinement. Although these two strategies for feature extraction and feature refinement employed in SVS-net explicitly increase the parameter number, they are necessary as the verification in the hyper-parameter experiments in Sec. \ref{sec2.2}. Moreover, the relatively long inference time mainly results from the feature fusion and channel attention mechanism. In the future work, we intend to explore more efficient network architectures for further decreasing computation time and improving inference efficiency.

\subsection{Downstream works}
Vessel segmentation is an efficient preprocessing procedure for various medical tasks. To assess the influence of vessel segmentation on various medical tasks, we further investigate two down-stream tasks that use the proposed SVS-net. We choose three state-of-the-art segmentation methods (\textit{i.e.}, SU-Net, BTS-DSN, X-ray net) to compare our SVS-net.

Estimating the distribution of coronary vessel networks via vessel segmentation is very important to evaluate the coronary circulation \citep{vigneshwaran2019reconstruction} in percutaneous coronary intervention. Usually, we estimate the area proportion of vessel network distribution in the whole heart regions of XCA images. Obviously, the wider the distribution, the smoother the blood flows in coronary circulation. Specifically, we use relative gland volume errors (GVEs) defined in \citep{Ghavami2019AutomaticSO} to measure the vessel distrubution area. GVE is calculated by the absolute difference between the predicting segmentation $V\left( {{y_{pr}}} \right)$  and the manual ground-truth segmentation $V\left( {{y_{gt}}} \right)$:
\begin{equation}\label{GVE1}
\text{GVE} = \frac{{\left| {V\left( {{y_{gt}}} \right) - V\left( {{y_{pr}}} \right)} \right|}}{{V\left( {{y_{gt}}} \right)}} \times 100\%
\end{equation}

$V\left(*\right)$ is based on counting the positive voxels in the binary segmentation. From the definition, it is easy to learn that a good segmentation method should have low GVE value. The relative GVE is summarized in Table \ref{GVE}. From Table \ref{GVE}, we can see that the mean GVE of SVS-net is $9.74\%$, which is much lower than those of other methods. Besides, the standard deviation of our method is also lower than those of other methods. These measures implicate that SVS-net is better and more stable in vessel distribution estimation than other methods. 

\begin{table}
	\caption{Vessel network volume calculations between the manual and segmentation of each method.}
	\label{GVE}       
	\begin{tabular}{ p{5cm} | p{1.4cm} p{1.4cm} p{1.4cm}  p{1.4cm}}
		\hline\noalign{\smallskip}
		Method & X-ray net & SU-Net & BTS-DSN & SVS-net \\
		\noalign{\smallskip}\hline\noalign{\smallskip}
		Relative GVE difference (\%) mean $\pm$ std &13.00 $\pm$10.35 & 23.25 $\pm$12.30 & 17.41 $\pm$11.47 & \textbf{9.74 $\pm$7.52} \\
		\noalign{\smallskip}\hline
	\end{tabular}
\end{table}
\begin{figure*}
	\centering
	\includegraphics[width=0.8\linewidth]{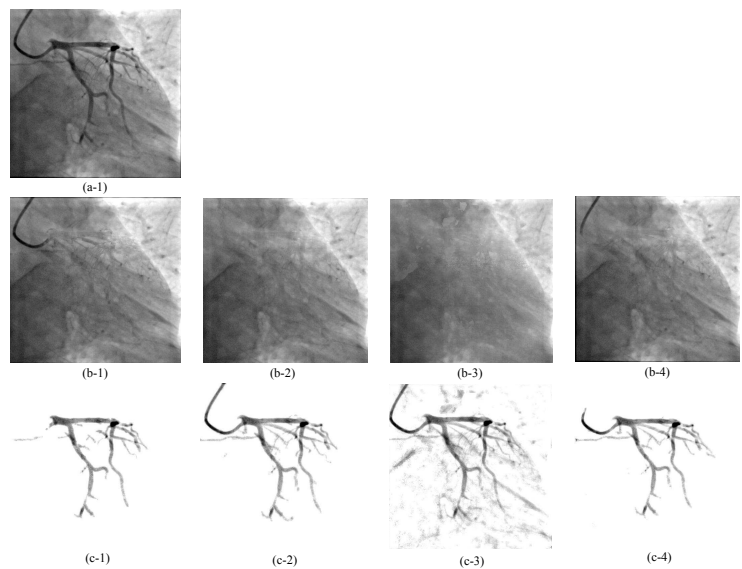}
	\caption{Vessel gray intensity recovery. From top to bottom, each row displays the original XCA image, the background and foreground vessel images recovered from SU-Net, X-ray net, BTS-DSN and SVS-net, respectively. }
	\label{Fig11}
\end{figure*}

Furthermore, quantitative coronary analysis and corresponding myocardial perfusion analysis are another downstream works for the diagnose and therapy of coronary artery disease. The gray‐level intensities of cardiac vessels carry important informations for the quantitative analysis. Usually, we locate the vessel through segmentation but lose the gray intensity at the same time. As illustrated in \citep{QIN201938}, we can use the vessel mask regions from the vessel segmentation and then complete vessel gray information in these regions by tensor completion algorithm \citep{QIN201938}. We compare three segmentation methods' effect on the final gray intensity recovery after vessel mask region segmentation. As shown in Fig \ref{Fig11}, SVS-net is much more conducive to reconstruct vessels and their gray intensities from the complex and noisy backgrounds of X-ray images. There are few vessel residuals remained in the background regions. Hence, SVS-net can provide relatively accurate vessel mask segmentation for recovering gray intensity in quantitative coronary analysis. 

\section{Discussion and Conclusion}
We propose a sequential vessel segmentation deep network, which integrates 3D convolutional layers extracting rich temporal-spatial features and utilizes CAB learning discriminative features from the complex and noisy background artifacts in the XCA image sequences. Experiment results verify the superior performance of special designs in our SVS-net. The proposed SVS-net can effectively segment the whole branches of the vessel trees from the XCA sequences. There is still room in the future work to improve the accuracy of segmenting small branches of thin vessels and enhance temporal-spatial consistency of vessel tree mask. To achieve a reliable segmentation of all small and thin vessels in the low-contrast and noisy XCA sequence, the channel-wise attention scheme can be further integrated into pixe-wise (or superpixel wise) saliency-aware image matching \citep{qin2018joint,qin2016structure} and segmentation \citep{wang2018saliency} methods to automatically choose the key frame that contains the most salient small and thin vessels from the XCA sequence so that this frame's feature representation can be taken as priors for pixel-wise labeling in sequential vessel segmentation. Deep feature matching \citep{Kong2018CrossDomainIM} and deep temporal-spatial correlation \citep{wang2019face} in the image sequence can also be utilized to transfer the learning priors from key frame to its neighbouring frames containing unsharp small vessels. 

Furthermore, we can sample vessel mask regions using contrast agent motion information or randomly sample vessel mask regions for background inpainting via tensor (or matrix) completion \citep{QIN201938,unberath2017consistency}, the completed background is then subtracted from XCA image sequence for the overall vessel extraction. This scheme of trial-and-completion can not only accurately recover the structures and intensities of vessel trees but also well compensate the deficiency of small vessel extraction (or segmentation) in the XCA image sequences. Such vessel extraction can be effectively implemented in an unsupervised deep network \citep{sultana2019unsupervised}.

\section*{Acknowledgments}
	This work was partially supported by the National Natural Science Foundation of China (61271320, 81370041 and 81400261) and Shanghai Jiao Tong University Cross Research Fund for Translational Medicine (ZH2018ZDA19). Yueqi Zhu was partially supported by three-year plan program by Shanghai Shen Kang Hospital Development Center (16CR3043A). BF was partially supported by NIH Grants R01CA156775, R21CA176684, R01CA204254, and R01HL140325. The authors would like to thank all authors for opening source codes used in the experimental comparison in this work. The authors would also like to thank the anonymous reviewers whose contributions considerably improved the quality of this paper.

\bibliography{mybibfile}

\end{document}